\documentclass{article}

     \usepackage[preprint]{neurips_2025}

\usepackage[utf8]{inputenc} 
\usepackage[T1]{fontenc}    
\usepackage{hyperref}       
\usepackage{url}            
\usepackage{booktabs}       
\usepackage{amsfonts}       
\usepackage{nicefrac}       
\usepackage{microtype}      
\usepackage{xcolor}         
\usepackage[normalem]{ulem}
\usepackage{multirow}
\usepackage{pifont}
\usepackage{fontawesome5}
\usepackage{tikz}
\usepackage{array}
\usepackage{xspace}
\usepackage{wrapfig}
\usepackage{enumitem}
\usepackage{threeparttable}
\usepackage{color-edits}
\addauthor{soni}{orange}
\addauthor{vc}{blue}
\addauthor{gn}{magenta}
\addauthor{xw}{violet}

\title{Coding Agents with Multimodal Browsing are Generalist Problem Solvers}

\author{%
  \textbf{Aditya Bharat Soni$^1$} \quad
  \textbf{Boxuan Li$^2$}  \quad 
  \textbf{Xingyao Wang$^3$}  \quad 
  \textbf{Valerie Chen$^1$}  \quad 
  \textbf{Graham Neubig$^{1,3}$} \\ \\
  $^1$Carnegie Mellon University \ \
$^2$Independent \ \
$^3$All Hands AI \ \
 \\
  \texttt{\{adityabs, gneubig\}@cs.cmu.edu}\\
}

\usepackage{pdfpages}
\begin{document}


\maketitle

\begin{abstract}
Modern human labor is characterized by specialization; we train for years and develop particular tools that allow us to perform well across a variety of tasks.
In addition, AI agents have been specialized for domains such as software engineering, web navigation, and workflow automation.
However, this results in agents that are good for one thing but fail to generalize beyond their intended scope.
One reason for this is that agent developers provide a highly specialized set of tools or make architectural decisions optimized for a specific use case or benchmark.
In this work, we ask the question: \emph{what is the minimal set of general tools that can be used to achieve high performance across a diverse set of tasks?}
Our answer is \AgentName, a generalist agent built with a modest number of general tools: code editing and execution, web search, as well as multimodal web browsing and file access.
Importantly, \AgentName demonstrates superior or competitive performance over leading specialized agents across three diverse and challenging benchmarks: SWE-Bench Multimodal \cite{yang2025swebench}, GAIA \cite{mialon2023gaia}, and The Agent Company \cite{xu2024theagentcompanybenchmarkingllmagents}, outperforming the best-performing previously published results with absolute improvements in success rate of \textbf{9.1}, \textbf{1.3}, and \textbf{9.1} points respectively. Further, we show how existing state-of-the-art multi-agent systems fail to generalize beyond their target domains. These results demonstrate the feasibility of developing a generalist agent to solve diverse tasks and establish \AgentName as a strong baseline for future research.\footnote{\AgentName is available open-source at: \url{https://github.com/adityasoni9998/OpenHands-Versa}}
\end{abstract}

\section{Introduction}\label{sec:introduction}

AI agents powered by Large Language Models hold great promise to accelerate or automate a wide variety of practical tasks.
For instance, agents have demonstrated strong software engineering capabilities and have been able to fix as many as two-thirds of issues in open-source Python repositories from SWE-Bench~\cite{jimenez2024swebench} and around one-third of issues in Javascript-based front-end libraries from SWE-Bench Multimodal~\cite{yang2025swebench}. 
In addition, agents have shown impressive web navigation capabilities, and can complete over half of the tasks from WebArena \cite{zhou2023webarena} and over one-third of the tasks from VisualWebArena~\cite{koh2024visualwebarena}. 
Agents have also exhibited strong performance as general assistants, solving over half of the tasks from GAIA~\cite{mialon2023gaia} that require various capabilities like gathering and synthesizing information from the web and processing multimodal data from diverse files. Finally, agents have also proven effective as digital workers, solving one-fourth of tasks in The Agent Company \cite{xu2024theagentcompanybenchmarkingllmagents} that require the agent to navigate company-internal websites and communicate with co-workers.

\begin{figure}[t]
    \centering\includegraphics[width=0.8\linewidth]{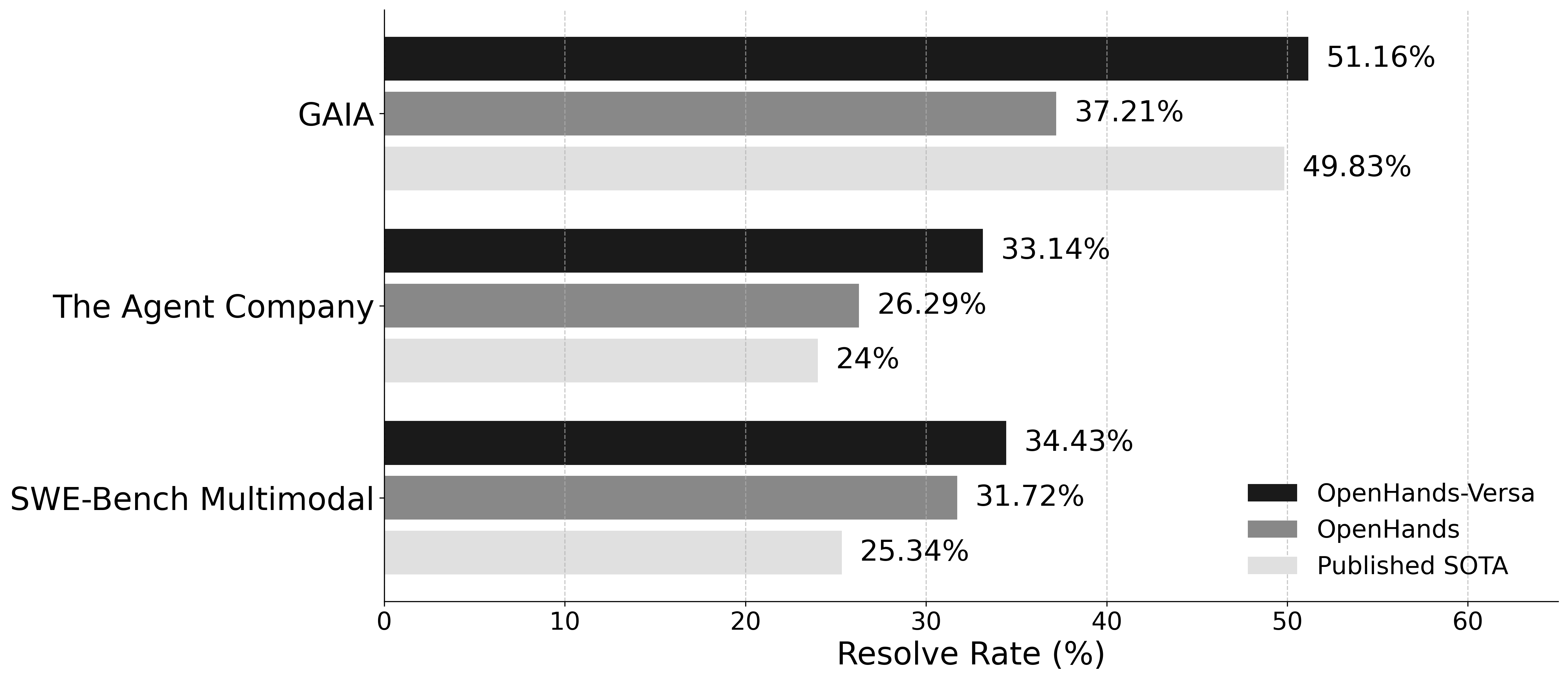}
    \vspace{-8pt}
    \caption{Comparison of \AgentName with previously published SOTA agents and OpenHands across GAIA, SWE-Bench Multimodal (SWE-Bench M) and The Agent Company. \AgentName outperforms the SOTA specialist agents for all three benchmarks. Notably, \AgentName improves browsing and information access abilities of OpenHands, while maintaining its software engineering capabilities. We focus on comparing to prior agents with reproducible code and results (more details in Table \ref{tab:results})}.  
    \label{fig:main_results}
    \vspace{-5pt}
\end{figure}

However, until this point, the strongest published agents in each domain have typically been explicitly optimized to perform well on a relatively narrow set of tasks and benchmarks, which we refer to as \textit{specialist agents}. 
For example, agents such as Agentless \cite{agentless} and SWE-Agent \cite{yang2024sweagent} have achieved state-of-the-art performance on SWE-Bench Python programming problems. Still, they cannot typically gather information from the web or communicate with co-workers via web-based chat platforms, which would result in poor performance on GAIA and The Agent Company, respectively. 
In contrast, strong web navigation agents like AgentSymbiotic \cite{zhang2025symbiotic}, AgentOccam \cite{yang2024agentoccam}, and Agent Workflow Memory \cite{wang2024agent} cannot write, debug, or execute code, so an agent with strong performance on WebArena may not be proficient on software engineering problems (e.g., SWE-Bench). More concretely, we refer to such agents as \textit{specialist agents} that either lack one or more capabilities, such as code execution, browsing, file viewing, search APIs, and have been primarily evaluated on a narrow category of tasks (i.e., one of browsing, coding, general assistance benchmarks).


Does this mean that we are destined for a world where each user must interact with a broad variety of agents, each specialized for a single task? In this paper, we argue that the answer to this is \emph{no}.
How could this be the case? We argue that a great majority of tasks can be tackled by agents that have the below three capabilities.
\begin{itemize}[noitemsep,topsep=0pt,parsep=0pt,partopsep=0pt,leftmargin=18pt]

\item \textbf{Coding:} The ability to write, debug, and execute code, including the use of libraries that are available to programmers.

\item \textbf{Multimodal Web Browsing:} The ability to browse the web, perform interactive actions (e.g., click, type) on webpages, and process vision and text modalities from webpages.

\item \textbf{Information Access:} The ability to search information on the web, typically using search APIs, and process multimodal content from various files such as PDFs, spreadsheets, code files, etc.
\end{itemize}

To implement such an agent, we propose \textbf{\AgentName}, built using the popular OpenHands framework~\cite{wang2024openhands} for coding agents, imbuing it with the ability to perform visual browsing, accessing information from the web through search APIs, and processing multimodal content from diverse files.

This simple strategy is surprisingly effective---we show a single agent can achieve strong results, rivaling or exceeding the state-of-the-art published systems, on three diverse and practical benchmarks: GAIA for general assistance and information access \cite{mialon2023gaia}, The Agent Company for evaluating agents as digital co-workers in a company~\cite{xu2024theagentcompanybenchmarkingllmagents}, and SWE-Bench Multimodal for frontend-focused software engineering \cite{yang2025swebench}, as shown in Figure \ref{fig:main_results}. Notably, \AgentName achieves state-of-the-art performance on all three benchmarks with absolute improvement of \textbf{9.1}, \textbf{1.3}, and \textbf{9.1} points in success rate over best-performing previously published results on SWE-Bench Multimodal, GAIA and The Agent Company respectively. Furthermore, \AgentName improves the browsing and information access abilities of OpenHands while retaining its coding capabilities. Notably, we find that current state-of-the-art multi-agent systems fail to generalize beyond their intended scope.

We also study the tool-use patterns of \AgentName and provide insights into why this simple strategy works so well. We find that \AgentName uses appropriate tools that align well with task requirements and has better domain-aware tool-selection than OpenHands. We also perform extensive analysis of the results and find some highly complex tasks that can be solved by \AgentName, while also pointing out error behaviors of our agent that can be addressed by future work.

\section{Towards a Generalist Agent}

\subsection{Preliminaries}\label{sec:openhands_basic_intro}
To implement a generalist agent, 
we choose to build \AgentName on top of the OpenHands (\cite{wang2024openhands}) framework.\footnote{We use OpenHands v0.28.1.}
OpenHands offers a flexible event stream architecture, a sandboxed runtime, a built-in evaluation harness for evaluating agents on numerous benchmarks, and the following tools:

\begin{itemize}[noitemsep,topsep=0pt,parsep=0pt,partopsep=0pt,leftmargin=18pt]
    \item[1.] A bash shell that connects to the operating system environment and supports the execution of Unix-style commands.
    \item[2.] Interactive python code execution via a Jupyter IPython server.
    \item[3.] A text-based browsing tool that uses a Chromium browser based on Playwright\footnote{https://playwright.dev/python/} and uses the BrowserGym framework (\cite{chezelles2025the}) to implement its action space.
    \item[4.] A file-processing tool for creating, viewing and editing plain-text files (e.g., files with extensions like .py, .txt, .cpp, .js, .json etc.).
     
\end{itemize}

\subsection{Ingredients of Our Agent}\label{sec:our_method}
Since OpenHands has primarily been an agent developed for software engineering, with strong \textbf{coding abilities} and support for multiple programming languages~\cite{zan2025multiswebench},
it lacks other capabilities like multimodal web browsing and information access. 
We augment OpenHands with these capabilities while inheriting the coding capabilities from OpenHands. 

\textbf{Multimodal Web Browsing:} The browsing tool in OpenHands relies solely on text-based observations that represent web pages using its accessibility tree (AXTree)\footnote{https://developer.mozilla.org/en-US/docs/Glossary/Accessibility\_tree}, and misses critical visual cues from the frontend. Instead, we adopt the Set-of-Marks prompting method \cite{yang2023setofmark}, which captures a screenshot of the current viewport (the visible area of a webpage in the browser window), overlays bounding boxes on interactable elements (e.g., buttons, links, text boxes), and labels them with unique alphanumeric browsergym-ids \cite{chezelles2025the} (e.g., refer to Appendix~\ref{app:som_example}). 
Note that this is similar to BrowserGym's GenericAgent \cite{chezelles2025the}.
We also include the full AXTree to provide context beyond the viewport but truncate to the current viewport if the AXTree is too large for the LLM's context window.

We also incorporate context condensation into the browser tool. 
OpenHands uses an event stream architecture wherein the backbone LLM is provided with action-observation pairs from all the previous steps at the next execution step. Since each browser observation consists of a webpage screenshot and a possibly large AXTree, this approach results in high inference costs, large observations that do not fit in the LLM's context window, and increases the agent's runtime due to slower LLM inference. 
To address this, we implement a browsing condenser that retains only the $k$ most recent browsing observations and masks out each older browsing observation with a fixed placeholder message. 

\textbf{Information Access:} We discuss two ways we improve information access: \textit{web search} and \textit{multimodal file processing}. First, synthesizing information from the web using search engines is crucial. While agents can achieve this by opening search engines like Google or Bing in the browser, in practice, we observe that the agent is frequently blocked by CAPTCHAs~\cite{eurocrypt-2003-2005}. We mitigate this by allowing the agent to perform a web search using a search API. 
This has the added advantage of reduced costs over browsing the web when searching for factual information and allows the use of specialized search APIs designed for agents. We use the Tavily API~\cite{tavily} for most of our experiments; however, \AgentName also supports the use of Exa~\cite{exa} and Brave~\cite{brave} APIs. 

Second, several tasks require the agent to access information about multimodal data from various files such as PDFs, audio files, presentation slides, etc. 
However, OpenHands has a limited file viewing support restricted to files that can be opened in text editors (like those with .txt, .py, .json, .js extensions).
We enhance the existing file viewing functionality of OpenHands by integrating Markdown converters, similar to those used by the FileSurfer in Magentic-One (\cite{fourney2024magentic}), to transform various files into a unified Markdown representation.

\textbf{Task Planning:} Task planning is crucial for multi-step execution, where agents must decompose the end-goal into multiple sub-tasks and organize their actions into a logical sequence. Prior approaches include developing an orchestrator or a planning agent \cite{fourney2024magentic,bahdanau2024tapeagents}, a Think tool \cite{claudethinktool} that the agent can flexibly invoke to log its plan, and using Chain-of-Thought prompting \cite{chezelles2025the}. We rather use a simple approach of appending a fixed planning prompt to the agent's event stream after every $\tau$ steps asking the agent to summarize its current progress and create a plan for the subsequent task execution. 

\subsection{Comparison with Existing Agents}\label{sec:compare_agent}
Next, we compare \AgentName with existing agents that excel in specific domains and benchmarks.
For coding agents, we consider SWE-Agent~\cite{yang2024sweagent} and Agentless~\cite{agentless}.
BrowserGym GenericAgent~\cite{chezelles2025the} and Browser-use~\cite{browser_use2024} are well-suited for web navigation. Multi-agent frameworks like Magentic-One~\cite{fourney2024magentic}, OpenDeepResearch~\cite{opendeepresearch}, and OWL-roleplaying~\cite{owl2025} excel at general-purpose assistance by synthesizing information from the web and processing various files. We also include OpenHands in this comparison.  Note that we only consider agents with open-source implementations, since this comparison requires understanding of their internal design.

\begin{table}[t]
    \caption{Comparison of different agents based on their tool-use capabilities. Definitions for the symbols - \iconCodeExecution : supports executing code, \iconFileEdit : supports editing operations, \iconBrowseInteractive : uses textual browsing, \iconScreenshot : uses visual browsing, \iconSearch : supports API-based search, \iconMultimodal : supports viewing multimodal file content, \iconTextOnly: supports viewing only plain-text files, \iconNo : capability not supported, \iconSingleAgent : uses a single agent framework, \iconMultiAgent : uses a multi-agent framework.}
    \centering
    \begin{tabular}{>{\bfseries}l c c c c c}
        \toprule
        \textbf{Method} & \textbf{Coding} & \textbf{Browsing} & \textbf{Search} & \textbf{File Viewing} & \textbf{Agents} \\
        \midrule
        SWE-Agent \cite{yang2024sweagent} & \iconCodeExecution \iconFileEdit & \iconNo & \iconNo & \iconTextOnly & \iconSingleAgent \\
        Agentless \cite{agentless} & \iconFileEdit & \iconNo & \iconNo & \iconTextOnly & \iconSingleAgent \\
        OpenHands \cite{wang2024openhands} & \iconCodeExecution \iconFileEdit & \iconBrowseInteractive & \iconNo & \iconTextOnly & \iconSingleAgent \\
        BrowserGym GenericAgent \cite{chezelles2025the} & \iconNo & \iconScreenshot & \iconNo & \iconNo & \iconSingleAgent \\
        Browser-use \cite{browser_use2024} & \iconNo & \iconScreenshot & \iconNo & \iconNo & \iconSingleAgent \\
        OpenDeepResearch \cite{opendeepresearch} & \iconCodeExecution & \iconBrowseInteractive & \iconSearch & \iconMultimodal & \iconMultiAgent \\
        OWL-roleplaying \cite{owl2025} & \iconCodeExecution & \iconScreenshot & \iconSearch & \iconMultimodal & \iconMultiAgent \\
        Magnetic One \cite{fourney2024magentic} & \iconCodeExecution & \iconScreenshot & \iconSearch & \iconMultimodal & \iconMultiAgent \\
        \AgentName (Ours) & \iconCodeExecution \iconFileEdit & \iconScreenshot & \iconSearch & \iconMultimodal & \iconSingleAgent \\
        \bottomrule
    \end{tabular}
\label{tab:agent_contrast}
\end{table}

The comparison of these agents with \AgentName is mainly along three core capabilities defined in \S\ref{sec:introduction}. Also, we examine whether the system adopts a multi-agent framework with specialized agents for distinct skills (such as web navigation, coding, and planning), or a single-agent framework, where one agent utilizes all available tools to complete the task. Table \ref{tab:agent_contrast} captures nuanced differences between these agents, which are described below. 

\textbf{Coding:} For software engineering (SWE) tasks, agents must have the ability to write and execute code, debug code by editing files, and use a shell to run tests, install packages, and navigate the repository. Browser-use and BrowserGym GenericAgent are designed exclusively for web navigation and lack all code-related abilites. Multi-agent systems like OWL-roleplaying, OpenDeepResearch, and Magentic-One support a subset of the these abilities but they lack support for editing files, implying that the agent has to regenerate the entire code from scratch when making any changes to existing files. Also OWL-roleplaying and OpenDeepResearch do not have access to a shell. These multi-agent systems mainly support the execution of stand-alone Python programs, making them unsuitable for SWE tasks and coding in other programming languages. Agents like SWE-Agent, OpenHands, and \AgentName support all the above code-related capabilities and are well-suited for SWE tasks. Although Agentless does not have a bash terminal, it supports the execution of selected tests in the repository to validate candidate patches within a human-defined workflow.

\textbf{Web Browsing: } Agents should be able to browse the web and execute interactive actions on websites to handle tasks such as filling online forms, ordering items from e-commerce websites and reading software documentation. Agents must also have a strong multimodal processing ability to comprehend the webpage content by jointly interpreting the visual layout (i.e., the rendered UI elements) and the webpage text. SWE-agent and Agentless do not support browsing, making them unsuitable for many practical tasks. OpenDeepResearch has a very limited browsing ability, wherein it can only open and scroll through webpages, without the ability to execute any other actions like ``click'' or ``type''. OpenHands supports interactive browsing actions, but it performs text-only browsing whereas all other agents perform mulimodal web browsing using visual context from webpage screenshots. 

\textbf{Information Access via Web Search: } Agents must be able to query search engines using keywords to retrieve relevant URLs, synthesize factual information, and access up-to-date content. Search APIs provide a more robust mechanism for supporting this functionality and mitigate issues caused by access restrictions like CAPTCHAs. Despite this tool being useful for several real-world tasks, most existing agents do not have this ability except multi-agent systems that demonstrate strong performance on the GAIA benchmark (\cite{mialon2023gaia}). This provides further evidence that many agent designs are over-tailored to specific domains or benchmarks.

\textbf{Information Access via Multimodal File Processing: } Agents must be able to view the content of various file types such as PDFs, presentation slides, spreadsheets etc. Although the agent can also read certain files using code or shell, this approach is prone to bugs and may require multiple attempts for successfully parsing the file. Supporting file viewing as a tool is a more robust approach since the agent can access content of various files through a single tool call. Web agents like Browser-Use and BrowserGym Generic Agent do not support file viewing. SWE-Agent, Agentless, and OpenHands have limited file-viewing support wherein the agent can only read plain-text files. All other agents have specific tool(s) that allow the agent to process multimodal file content.
\section{Experimental Setup}\label{sec:experiments}
In this section, we describe our experimental setup to demonstrate the effectiveness of \AgentName. We overview our choice of evaluation benchmarks and the corresponding evaluation metrics in \S\ref{sec:benchmarks}, and discuss our baselines in \S\ref{sec:baselines}.

\subsection{Evaluation Benchmarks}\label{sec:benchmarks}
We evaluate \AgentName on three benchmarks that cover a diverse range of capabilities and agent use cases---which can be roughly gleaned from Figure~\ref{fig:tool_use_freq}. 

\textbf{SWE-Bench Multimodal (SWE-Bench M)~\cite{yang2025swebench} : } This benchmark evaluates the ability of agents to fix software issues in GitHub repositories of front-end, user-facing libraries. 
The benchmark requires the agent to solve GitHub issues from 17 popular JavaScript code repositories for various use-cases like web development, syntax highlighting, and data visualization. 
Furthermore, several tasks also have visual assets (images and videos) describing the issue and links to online integrated development environments (IDEs) containing code snippets for reproducing the issue, requiring agents to process multi-modal data to visually analyze the issue. 
Unlike SWE-Bench (\cite{yang2025swebench}), where all the reference solutions 
only require editing Python files, more than 28\% of SWE-Bench M instances require editing files across two or more programming languages.

\textbf{GAIA~\cite{mialon2023gaia}:} This benchmark evaluates AI agents as general-purpose assistants using tasks that require browsing the open web, performing web search, coding, reasoning, and processing multimodal content from audios, spreadsheets, and PDFs. While the coding tasks in SWE-Bench M require agents to fix issues by editing \emph{existing} code files, the coding tasks in GAIA generally require the agent to write and execute stand-alone programs from scratch. 

\textbf{The Agent Company \cite{xu2024theagentcompanybenchmarkingllmagents}:} This benchmark evaluates the ability of agents as digital co-workers in a simulated software company using a reproducible, self-hosted environment. It covers tasks across software development, project management, data science, financial analysis, etc. It uses four self-hosted websites: GitLab for hosting code repositories and documentation, OwnCloud for cloud-based file-sharing, Plane for issue tracking and project management, and RocketChat for communicating with simulated co-workers. To solve the tasks in this benchmark, the agent must be able to browse websites, write code, communicate with simulated colleagues, and read, write and edit various files.

\subsection{Baseline Agents}\label{sec:baselines}
For each benchmark, we compare to the best-performing open source agent frameworks (from the benchmark's leaderboard) that have reproducibility guidelines \footnote{On the GAIA leaderboard there are other methods with no code or technical details published with scores of up to 77\%. In this paper, we focus on methods that have available details and reproducible code bases.}.

For \textbf{SWE-Bench Multimodal}, we choose Agentless-Lite \cite{dunn2025agentlesslite}, and SWE-Agent \cite{yang2024sweagent} along with its Multimodal and Javascript variants proposed along with this benchmark. For \textbf{GAIA}, we consider Magentic-One \cite{fourney2024magentic} and OpenDeepResearch \cite{opendeepresearch}. 
For \textbf{The Agent Company}, we consider OWL-roleplaying \cite{owl2025} and OpenHands v0.14.2, which is the version used in the original paper.
Across all benchmarks, we evaluate OpenHands v0.28.1---the agent on top of which \AgentName is built---to understand the effect of our modifications. 

Most baseline agents report performance on only one of the benchmarks, and their architecture typically does not support evaluation on the others (as discussed in Section~\ref{sec:compare_agent}). 
Agentless-lite and all SWE-agent variants cannot browse, use search engines, or process multimodal files, which are crucial for GAIA and The Agent Company. 
Multi-agent baselines for GAIA cannot typically write code in languages other than Python (like JavaScript), making them unsuitable for SWE-Bench M(\S\ref{sec:our_method}).

We used \texttt{claude-3-7-sonnet-20250219} as the backbone LLM of \AgentName and OpenHands v0.28.1 to ensure a direct comparison between the two agent architectures using the same LLM. We also evaluate OpenHands-Versa with the recently released \texttt{claude-sonnet-4-20250514} model. Further experimental details are provided in \S\ref{sec:agent_config}.

\subsection{Evaluation metrics}
For all the benchmarks, we use the evaluation metrics proposed by the authors of the corresponding benchmarks. We use resolve rate for GAIA and SWE-bench M -- the \% of tasks completed successfully by the agent. The Agent Company has checkpoint-based evaluation wherein two metrics are computed: \textbf{Full completion score} (the \% of tasks for which all checkpoints were resolved) and \textbf{Partial completion score} (also provides partial credit for successful checkpoints in partially completed tasks). We refer the reader to The Agent Company \cite{xu2024theagentcompanybenchmarkingllmagents} for more details. We use the test split for all three benchmarks (The Agent Company does not have a validation split). In addition, we report \textbf{ pass @ 1} metrics for all benchmarks.

\section{Main Results}\label{sec:results}

\begin{table}[t]
\centering
\caption{Comparison of agent performance across GAIA, SWE-bench M, and The Agent Company. Highest metrics for each benchmark are \textbf{bold-faced} and second highest metrics are \underline{underlined}. When available, we report the metrics directly as mentioned by the baseline agents on the respective benchmark leaderboards. Note that we restrict our comparison to agents with open-source implementations and method description, and exclude systems that either do not have guidelines to reproduce their performance or do not provide details about their exact agent configuration.}
\renewcommand{\arraystretch}{1.2}
\resizebox{\textwidth}{!}{%
\begin{tabular}{lc|ccccc}
\toprule
\textbf{Agent} & \textbf{Model(s)} & \textbf{GAIA} & \textbf{SWE-bench M} & \multicolumn{2}{c}{\textbf{The Agent Company}} \\
\cmidrule(lr){5-6}
 &  &  &  & \textbf{Full} & \textbf{Partial} \\
\midrule
Magentic-One \cite{fourney2024magentic} & gpt-4o, o1 & 37.87\% & - & - & - \\
\addlinespace
OpenDeepResearch \cite{opendeepresearch} & o1 & \underline{49.83\%} & - & - & - \\
\addlinespace
\midrule
SWE-Agent \cite{yang2024sweagent} & gpt-4o & - & 11.99\% & - & - \\
 & claude-3.5 sonnet & - & 12.19\% & - & - \\
\addlinespace
SWE-Agent JS \cite{yang2025swebench} & gpt-4o & - & 9.28\% & - & - \\
 & claude-3.5 sonnet & - & 11.99\% & - & - \\
\addlinespace
SWE-Agent Multimodal \cite{yang2025swebench} & gpt-4o & - & 12.19\% & - & - \\
 & claude-3.5 sonnet & - & 11.41\% & - & - \\
\addlinespace
Agentless-Lite \cite{dunn2025agentlesslite} & claude-3.5 sonnet & - & 25.34\% & - & - \\ \midrule
OWL-roleplaying \cite{owl2025} & gpt-4o, o3-mini & - & - & 4.00\% & 11.04\% \\
OpenHands v0.14.2 \cite{wang2024openhands} & gpt-4o & - & - & 8.60\% & 16.70\% \\
 & gemini-2.0 flash & - & - & 11.40\% & 19.00\% \\
 & claude-3.5 sonnet & - & - & 24.00\% & 34.40\% \\ \midrule
\addlinespace
OpenHands v0.28.1 \cite{wang2024openhands} & claude-3.7 sonnet & 37.21\% & \underline{31.72\%} & 26.29\% & 36.41\% \\ \midrule
\addlinespace
\AgentName & claude-3.7 sonnet & \textbf{51.16\%} & 31.33\% & \underline{30.86\%} & \underline{40.18\%}\\
& claude-sonnet-4 & \textbf{51.16\%} & \textbf{34.43\%} & \textbf{33.14\%} & \textbf{43.19\%}
\\
\bottomrule
\end{tabular}}
\label{tab:results}
\end{table}


We present our experimental results in Table~\ref{tab:results} and highlight the key takeaways below.

\textbf{\AgentName outperforms or matches existing agents for all three benchmarks: }Notably, \AgentName achieves state-of-the-art or close to state-of-the-art performance on all three benchmarks with \textbf{51.16\%} resolve rate on GAIA, \textbf{34.43\%} resolve rate on SWE-Bench M, and \textbf{33.14\%} full completion score on The Agent Company with claude-sonnet-4 as the backbone LLM. \AgentName outperforms the strongest baseline for GAIA with an absolute improvement of \textbf{1.33} points. On GAIA, \AgentName outperforms complex, multi-agent systems which use specially designed agents for distinct skills/sub-tasks, with each agent using a separate LLM. In addition, \AgentName achieves state-of-the-art performance on The Agent Company with an absolute improvement of \textbf{6.9} points in the full completion score and \textbf{6.8} points in the partial completion score over the best-performing baseline. On SWE-Bench M, \AgentName demonstrates an absolute improvement in resolve rate of \textbf{9.09} points over Agentless-Lite and more than \textbf{22} points over SWE-Agent and its variants. Notably, these gains are achieved without specific optimizations for SWE-Bench M, such as the JavaScript linter in SWE-Agent JS and SWE-Agent Multimodal.

\textbf{\AgentName has stronger browsing and information access capabilities than OpenHands, while retaining its coding capabilities}: While attempting to improve the browsing and information access capabilities in \AgentName (\S\ref{sec:our_method}), it is also crucial to ensure that our changes do not cause regression in the coding abilities inherited from OpenHands. This is concretely validated by comparing the results of the two agents on the three evaluation benchmarks. When using the same backbone LLM (claude-3.7 sonnet), \AgentName significantly outperforms OpenHands on GAIA with an absolute improvement of \textbf{13.9} points in the resolve rate. Furthermore, for The Agent Company, \AgentName achieves an absolute improvement of \textbf{4.6} points in the full completion score and \textbf{3.8} points in the partial completion score over OpenHands. Furthermore, \AgentName achieves a nearly equal resolve rate on SWE-Bench M as that of OpenHands, with an absolute difference of only 0.39 points. 

\textbf{Multi-agent systems with strong performance on GAIA fail to generalize: }OWL-roleplaying is a complex multi-agent system with separate agents for browsing, planning, web search etc. It is one of the top performing agents on GAIA validation set with a 58.18\% resolve rate, but does not report performance on GAIA test set. However it fails to generalize to The Agent Company with a poor full completion score of \textbf{4\%} and partial completion score of \textbf{11.0\%}. OWL-roleplaying significantly underperforms \AgentName, with an absolute decrease in \textbf{29.1} points in the full completion score and \textbf{32.2} points in the partial completion score.
\section{What Went Right and What Went Wrong?}
In this section, we provide fine-grained analyses to understand the agent behaviour for different tasks. Since we observe that \AgentName outperforms or matches OpenHands on the three benchmarks, with a minimal set of changes, we first compare their tool use patterns (\S\ref{sec:tool_use}). Next, we perform a comprehensive error analysis of our agent, to understand its limitations and provide insights for future improvement (\S\ref{sec:error_analysis}). Finally, we also discuss the effect of the search API on downstream agent performance for GAIA (\S\ref{sec:search_engine_effect}).

\subsection{Tool Use Patterns across Benchmarks}\label{sec:tool_use}
\begin{figure}[h!]
        \centering
        \includegraphics[width=\linewidth]{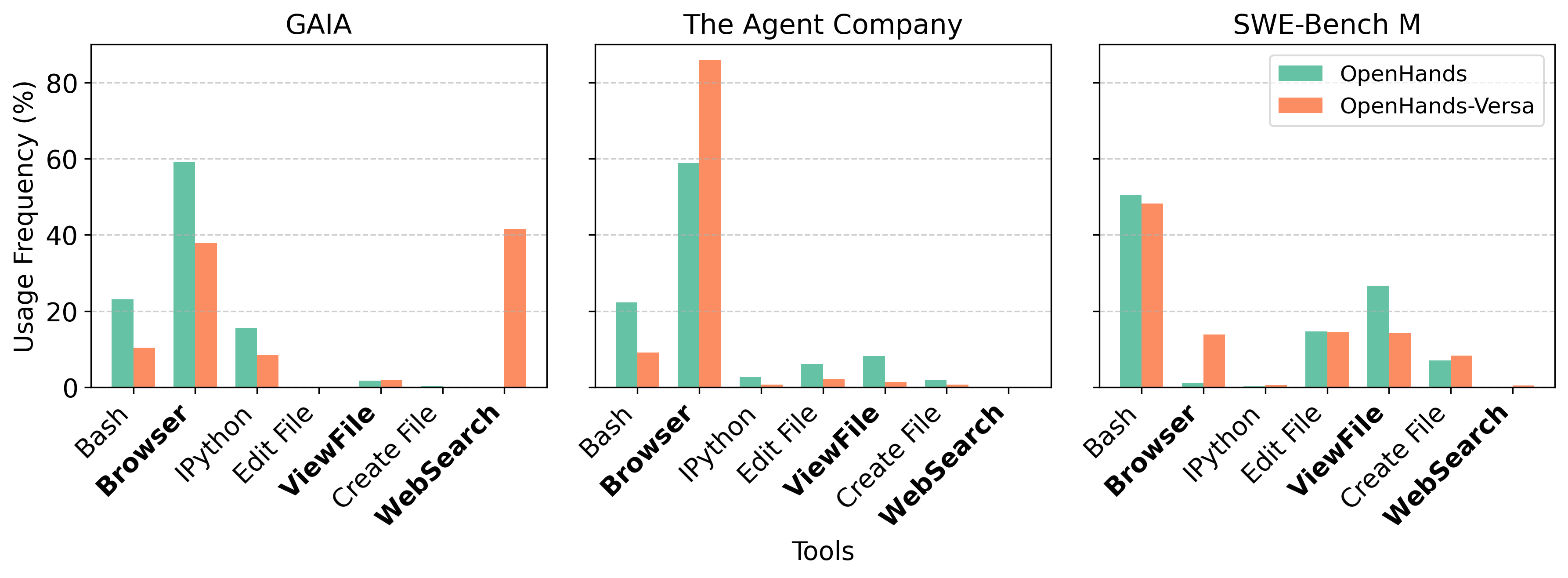}
        \caption{Distribution of the different tools used by OpenHands and \AgentName. \AgentName adapts its tool usage to different benchmarks without any benchmark-specific optimizations and \AgentName has better domain-aware usage of its tools as compared to OpenHands. Tools with \textbf{bold-faced} names have been modified/created by our work.}
        \label{fig:tool_use_freq}
\end{figure}

To better understand the behavior of \AgentName compared to the original OpenHands (when using the same LLM claude-3.7 sonnet), we plot the distribution of the relative tool use frequencies (as a percentage of total tool calls made by the agent) across all tasks, for all the 3 benchmarks in Figure \ref{fig:tool_use_freq}. This figure presents some interesting insights into the behavior of \AgentName and OpenHands, which we describe below: 

\textbf{\AgentName uses appropriate tools that align well with task requirements: } Our analysis shows that \AgentName generally selects tools that intuitively align with various tasks. For GAIA, the agent primarily uses the browser and search engine, consistent with the need to synthesize web-based information, and makes limited use of file-editing tools while frequently executing standalone Python code via IPython. It uses the bash tool in creative ways. For example, to install packages and download files with \texttt{wget} -- highlighting flexible problem-solving. In The Agent Company, tool usage is dominated by the browser, which reflects the benchmark’s focus on navigating internal websites, with little reliance on the search engine or IPython since URLs of company websites are known to the agent and tasks involve modifying code from repositories. For SWE-Bench M, the agent frequently uses bash, edit\_file, and view\_file tools, in line with practical software engineering workflows, and leverages the browser to visually verify its changes by rendering HTML files, demonstrating a nuanced understanding of front-end development practices.

\textbf{\AgentName has better domain-aware tool-selection than OpenHands: } For GAIA, OpenHands relies more heavily on the browser in the absence of a search engine, due to which it frequently navigates to hallucinated or invalid URLs. \AgentName first uses the search engine to retrieve relevant links and then chooses a URL based on the retrieved snippets, resulting in more targeted navigation. For SWE-Bench Multimodal, while overall tool usage patterns are similar, \AgentName makes more frequent use of the browser for visual verification of front-end changes, a capability that OpenHands cannot exploit due to text-only browsing. For The Agent Company, both agents display similar tool-use behavior, which is expected since changes to the browser tool in \AgentName primarily improve browsing observations rather than changing or expanding its action space.

\subsection{Error Analysis}\label{sec:error_analysis}

Next, we manually analyse the trajectories of \AgentName, describe its error behaviors, and provide some examples in Table \ref{tab:error_behaviors}.
\begin{table}[t]
\centering
\caption{Example tasks for some observed error behaviors of \AgentName.}
\renewcommand{\arraystretch}{1.2}
\resizebox{\textwidth}{!}{%
\begin{tabular}{p{3cm}|p{9cm}|p{5cm}}
\toprule
\textbf{Benchmark} & \textbf{Task Description (irrelevant details truncated)} & \textbf{Observed behaviour} \\
\midrule
\multirow{2}{*}{GAIA} & The Latin root of the Yola word ``gimlie'' shares a spelling with a Spanish word. What is the Google translation of the source title for the 1994 example sentence for that word in the Collins Spanish-to-English dictionary online? & Agent cannot access Collins dictionary website due to CAPTCHAs. \\
\cmidrule(lr){2-3}
 & In April of 1977, who was the Prime Minister of the first place mentioned by name in the Book of Esther? & Agent relies on incorrect search engine summary when searching for the first place given in the book.\\
\midrule
\multirow{2}{*}{The Agent Company} & We are collecting employees' preferences on drinks. Please navigate to ownCloud and find drinks\_survey.pdf and tell 3 most popular drinks to HR manager via RocketChat. & Agent gets stuck in a loop and fails to find the file on ownCloud.\\
\cmidrule(lr){2-3}
 & In Plane there open issues in the JanusGraph project. I want you to add all ``In Progress'' issues to Gitlab. & Agent fails to copy all issues and exits after partial completion of task.\\
\midrule
\multirow{2}{*}{SWE-Bench M} & Happiness Support card needs preventWidows treatment in WordPress. Steps to reproduce ... What I expected ... What happened instead ... & Agent does not write tests to verify its fix and does not follow steps to reproduce the bug. \\
\cmidrule(lr){2-3}
 & WebGL: render buffers are not always created correctly. The issue is that when creating a retained-mode geometry... & Agent does not execute existing tests in the repository due to which its changes fail the Pass-to-Pass tests. \\
\bottomrule
\end{tabular}}
\label{tab:error_behaviors}
\end{table}

For GAIA, we use the validation split since the ground truth is not available for the test set. We find that \AgentName is sometimes over-reliant on the retrieved summaries/snippets from the webpage given by the search API and uses factually incorrect information. We also find that the agent cannot access some websites due to various security measures like CAPTCHAs. For tasks in the SWE-Bench M, we find that the agent frequently struggles at creating comprehensive tests to verify its code, and prematurely exits, assuming that its code is correct since its non-exhaustive tests pass. Sometimes, the agent does not execute tests given in the repository to verify if its changes did not break existing functionality. For The Agent Company, we find that the agent generally struggles when interacting with OwnCloud, and frequently gets stuck in loops. Furthermore, we find that the agent sometimes prematurely exits without satisfying all the task requirements for the more complex tasks. 

\subsection{Effect of Search API on GAIA}\label{sec:search_engine_effect}
Since 40\% of the tool calls made by \AgentName for GAIA are to query the search engine with minimal use of this tool for other 2 benchmarks, we study the effect of choosing different search APIs on the downstream performance for this benchmark. We evaluated \AgentName on the GAIA validation split using three search APIs: Brave, Exa, and Tavily.

Although seemingly unimportant, the choice of search API significantly impacts downstream performance. We observe considerable variations in the resolve rate with \textbf{56.96\%} when using Brave, \textbf{58.18\%} when using Exa and \textbf{64.24}\% when using Tavily APIs. Notably, switching from Brave to Tavily results in an absolute improvement of \textbf{7.28} points in resolve rate. Our analysis shows that the agent relies on search snippets to decide which webpages to open for obtaining information. Brave extracts these snippets from raw webpage text, while Exa and Tavily provide higher-quality LLM-generated summaries. These often eliminate the need to open webpages in the browser, reducing inference costs due to large browsing observations as compared to the compact search results.  However, reliance on these summaries occasionally introduces hallucinations when they contain inaccuracies. Tavily partially mitigates these by offering an LLM-generated answer per query, synthesized from all retrieved results, which tends to be more accurate than individual page summaries. This is also one of the primary reasons why using Tavily has a higher resolve rate than other APIs.

\section{Conclusion}
In this work, we propose \AgentName -- a simple and flexible agent that demonstrates strong performance across three benchmarks, namely, GAIA, SWE-Bench M and The Agent Company. Our comprehensive experimental results demonstrate the effectiveness of \AgentName in tasks across various domains and highlight that generalizability can be achieved using an intuitive agent design without developing specialized agent implementations over-optimized for a particular domain. More concretely, these results indicate that generalist agents can be designed simply by providing the necessary tools to the agent and leaving it for the backbone LLM to autonomously decide how to use these tools to solve the task. Our results also demonstrate why existing agents fail to generalize beyond their target domain. We elaborate on the limitations of our approach and provide some insights for future improvement in \S\ref{app:limitations}. In conclusion, \AgentName will serve as a strong baseline for future research on generalist agents. 

\section*{Acknowledgement}
We would like to thank Frank Xu and Engel Nyst for helping us fix some of the bugs the initial agent design. We would also like to thank Mingchen Zhuge, Shuyan Zhou, and Pranjal Aggarwal for the helpful discussions.


\bibliographystyle{plain}
\bibliography{references}

\newpage
\appendix
\section{Webpage Screenshot with Set-of-Marks Annotation}\label{app:som_example}
\begin{figure}[!htb]
    \centering
    \includegraphics[width=\linewidth]{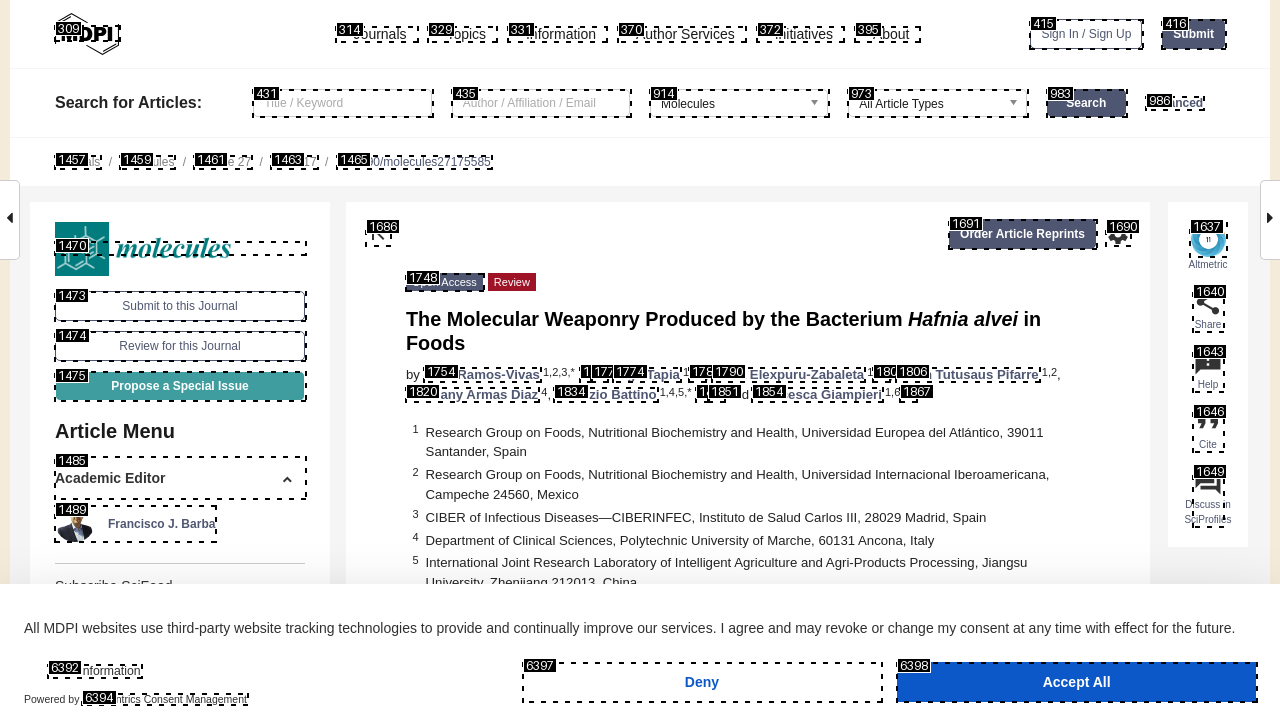}
    \caption{Example screenshot of a webpage with set-of-marks annotation}
    \label{fig:som_example_image}
\end{figure}

Figure \ref{fig:som_example_image} is an example screenshot of a webpage with all the interactable elements annotated with bounding boxes and their corresponding browergym-ids.

\section{Experimental Setup}\label{sec:agent_config}
In this section, we provide more details about our experimental setup. 

First, we discuss the exact configuration used for \AgentName. For browsing condensation (\S\ref{sec:our_method}), we set the context window ($k$) to 1 implying that we only retain the most recent browsing observation in the event stream. For planning, we set the planning interval ($\tau$) to 10, which implies that we append the planning prompt (\S\ref{sec:our_method}) to the event stream after every 10 steps. Notably, we use identical agent implementation for all the three benchmarks as opposed to other agents that selectively choose only relevant tools for different benchmarks, develop specialized tools to improve performance on a specific benchmark, or use benchmark-specific or domain-specific system prompts that will not generalize to all scenarios. For example, OWL-roleplaying provides tools to search Wikipedia and the Wayback Machine\footnote{\url{https://archive.org}}, which are particularly useful for GAIA since many tasks require the agent to search for factual information from Wikipedia and some tasks refer to websites that are no longer publicly available, requiring the agent to access them via the Wayback Machine. 

We set the temperature of the backbone LLM to 0 for all our experiments. We limit the maximum number of steps allowed for the agent to 100 for SWE-Bench M and The Agent Company, and to 60 for GAIA. Since GAIA requires the final answer given by the agent to exactly match with the ground truth answer, we extract the final answer of the agent using an LLM (particularly \texttt{claude-3-7-sonnet-20250219}) giving it the task description and the final thought of the agent. This also helps with some output formatting errors. For example, the agent may write the answer numerically (for eg. 500), whereas the task asks the agent to write it in text (i.e five hundred).

All our experiments are run using CPU-only, cloud-based machines (AWS EC2 instances -- t3.2xlarge specification with 32GB RAM, 8 vCPUs, and 512GB disk space). However, they can also run on local computers and do not require any GPU resources. The total runtime for evaluating \AgentName and OpenHands is $\approx$24 hours for GAIA, $\approx$54 hours for The Agent Company, and $\approx$12 hours for SWE-Bench M. Also, evaluating OWL on TAC takes  $\approx$50 hours.

\subsection{Baseline Agents}
Next, we provide more details about the baseline agents used in our work.

\textbf{Magentic-One\cite{fourney2024magentic}} is a generalist multi-agent system that uses an LLM-based Orchestrator Agent responsible for planning, tracking progress, and querying other agents/tools for different sub-tasks. Orchestrator can issue commands to WebSurfer, Coder, FileSurfer and ComputerTerminal. WebSurfer is an LLM-based agent responsible for web browsing and searching the web using Bing. Coder is an LLM-based agent that can write a new stand-alone Python program for each request and it should regenerate the entire code from scratch when debugging the code it previously wrote. The Orchestrator can read various files using the FileSurfer that converts different files in a unified Markdown format. Finally, the Orchestrator can run Unix-style commands in a shell using the  ComputerTerminal tool. This system does not have native support to create or edit files and write code in other programming languages.

\textbf{OpenDeepResearch \cite{opendeepresearch}} is a multi-agent system similar to Magentic-One. Its CodeAgent can write and execute stand-alone Python programs, read different files similar to FileSurfer in Magentic-One, ask questions about files, videos, and images to an LLM-based file viewer, and delegate browsing tasks to a separate browsing agent. The browsing agent uses a text-only browser to view webpages. It can only scroll on the webpage and search for text on a webpage, but cannot perform other actions like click, type, hover, etc. The browsing agent has tools to search the web using APIs and search the Wayback machine for archived webpages. The CodeAgent and the browsing agent each have their own planner agents that analyze their progress after every few steps and create a step-by-step plan. The CodeAgent is restricted to a fixed set of pre-installed libraries/packages that it can use. There is no native support for using a bash shell, writing and editing files, and executing code in other programming languages.

\textbf{OWL-roleplaying \cite{owl2025}} is a multi-agent system similar to Magentic-One and OpenDeepResearch. It has a user agent that assists with the task, creates plans, and issues commands to the assistant agent. The assistant agent is responsible for solving the task and has access to the various tools to extract content from different files, query LLMs to analyze images, videos and audios, execute stand-alone Python code, use an LLM-based search tool for searching the web using multiple search APIs, the WayBack machine, and Wikipedia, and delegate its browsing tasks to a separate browsing agent. The browsing agent has its own planner agent, uses visual browsing to browse the web, and can execute interactive actions on webpages. It has no native support for writing and editing files, using a bash shell, and executing code in other programming languages. Similar to OpenDeepResearch, it has a restrictive design wherein the agent can only use a fixed set of pre-installed libraries/packages for its Python programs.

\textbf{SWE-Agent \cite{yang2024sweagent}} is a software engineering agent that has access to a bash terminal, an agent-computer interface for reading, writing and editing code files, and a specialized Python-specific linter that checks if the edits made by the agent are syntactically correct. It cannot browse webpages, search the web, or read multimodal file content.

\textbf{SWE-Agent JS and SWE-Agent Multimodal \cite{yang2025swebench}} are extensions of SWE-Agent for the SWE-Bench M benchmark. SWE-Agent JS adds support for detecting errors in Javascript code edits made by the agent. SWE-Agent Multimodal is built on top of SWE-agent JS, and has the ability to serve local HTML code in a visual web browser, and open images. This allows the agent to visually reproduce image-based issues and visually verify its fixes. Just like SWE-Agent, none of these variants have the ability of browse public webpages, use search engines, or process multimodal file content.

\textbf{Agentless-Lite \cite{dunn2025agentlesslite}} is a lightweight version of Agentless \cite{agentless} that first uses RAG-based localization to retrieve the top 5 files that are relevant to the issue. Next it queries an LLM with these files to generate a patch. While it achieves impressive results with this simple method, its design is very limited. It does not support code execution, bash shell, multimodal file processing, web browsing, or using search engines.
\begin{table}[t]
\centering
\caption{Example tasks for each of the three benchmarks used in this work.}
\renewcommand{\arraystretch}{1.2}
\resizebox{\textwidth}{!}{%
\begin{tabular}{p{3cm}|p{9cm}|p{5cm}}
\toprule
\textbf{Benchmark} & \textbf{Task Description (irrelevant details truncated)} & \textbf{Capabilities/Tools required} \\
\midrule
\multirow{2}{*}{GAIA} & What animals that were mentioned in both Ilias Lagkouvardos's and Olga Tapia's papers on the alvei species of the genus named for Copenhagen outside the bibliographies were also present in the 2021 article cited on the alvei species' Wikipedia page about a multicenter, randomized, double-blind study? & Web search, Web Browsing, Multimodal file processing\\
\cmidrule(lr){2-3}
 & The attached image contains a Python script. Run the Python code against an array of strings, listed below. The output of the Python script will be a URL containing C++ source code. Compile and run this C++ code against the array [35, 12, 8, 99, 21, 5] and return the sum of the third and fifth integers in the sorted list. arr = [`\_alg', ..., `ht'] & Code execution, Multimodal file processing, Web Browsing.\\
\midrule
\multirow{2}{*}{The Agent Company} & We are collecting employees' preferences on drinks. Please navigate to ownCloud and find drinks\_survey.pdf and tell 3 most popular drinks to HR manager via RocketChat. & Web browsing, Multimodal file processing\\
\cmidrule(lr){2-3}
 & On our office cloud at http://the-agent-company.com:8092/, find the July-Sep 2024 financial report for our company, and create a SQLite database with two tables that appropriately populates the data in the report & Web browsing, Code Execution, Multimodal file processing\\
\midrule
\multirow{2}{*}{SWE-Bench M} & KML Symbol Align/Placement/Size. There is a bug with the anchor point for some symbols [Right Image] ...
I've attached a screen clipping from Google Earth to
show how it is supposed to look & Coding, Multimodal file processing (images and code files)\\
\cmidrule(lr){2-3}
 & Bracket highlighted with different color in class inheritance context.
- Reproduced in JSFiddle: https://jsfiddle.net/kkangmj/e7h48w36/7/
(Image) ... & Coding, Web Browsing, Multimodal file processing\\
\bottomrule
\end{tabular}}
\label{tab:examples}
\end{table}
\section{Performance on GAIA Validation Split}
We also evaluate \AgentName on the validation split of GAIA. Just like all other experiments, we consider agents with open-source implementation which have reproducibility guidelines and provide details about the exact configuration used by their agent.

Using the agent configuration described in \ref{sec:experiments} and claude-3.7-sonnet as the backbone LLM, \AgentName achieves a resolve rate of \textbf{64.24\%} on GAIA validation split. Notably, \AgentName outperforms top-performing, specialist, multi-agent systems -- Magentic-One (46.06\% resolve rate), OpenDeepResearch (55.15\% resolve rate) and OWL-roleplaying (58.18\% resolve rate).

\section{Example Tasks}

In this section, we provide some example tasks from each of the three benchmarks -- GAIA \cite{mialon2023gaia}, SWE-Bench M \cite{yang2025swebench}, and The Agent Company \cite{xu2024theagentcompanybenchmarkingllmagents}. Table \ref{tab:examples} shows some example tasks from each of the three benchmarks along with the tools or capabilities required to solve each of these tasks. Clearly, these examples qualitatively demonstrate that an agent must be proficient in several capabilities to perform well on all three benchmarks. Furthermore, they also help us understand why other agents will not be able to solve tasks from other benchmarks. In the absence of browsing, Agentless-Lite and SWE-Agent cannot solve any of the given examples for GAIA and The Agent Company. In the absence of Javascript code execution, none of the multi-agent systems can solve example tasks given for SWE-Bench M.

\section{Limitations of Our Approach}\label{app:limitations}
In this section, we describe the limitations of our proposed approach. Firstly, we do not consider tasks that involve interaction with GUI-based desktop computers like those in OSWorld \cite{xie2024osworldbenchmarkingmultimodalagents}. \AgentName and OpenHands both have access to a headless/non-GUI based operating system via the shell. Secondly, \AgentName has limited video processing abilities and cannot view local video files using the file viewer tool. Potential mitigations for this could be to use an LLM-based file summarizer/video summarizer. Thirdly, like most other agent frameworks, our work also primarily relies on closed-source LLMs. While it is feasible to use open-source LLMs for \AgentName, we observe that most AI agents perform very poorly when open-source LLMs are used. Finally, due to high cost of using proprietary LLMs, we are unable to evaluate every baseline agent on all the 3 benchmarks, but limit ourselves to strong baseline agents in order to empirically validate our hypothesis.

\section{Broader Impact of Our Work}\label{app:impact}
AI agents have shown promise in addressing complex tasks, but still face significant limitations when given long-horizon, real-world tasks. Our research advances the field by enhancing the generalizability of these systems and improving their performance across diverse practical applications. The work establishes a robust foundation for future developments in AI agent capabilities, particularly in the domain of generalizable agents.
However, these advancements bring important societal considerations. As AI agents become more sophisticated, potential risks emerge, including:
\begin{itemize}
    \item Misuse of agents for illegal or harmful activities like sending phishing emails, posting harmful content on social media, spreading misinformation etc.  
    \item Potential for causing large-scale unemployment and labor market disruption as agents become capable of performing tasks currently done by humans.
    \item Challenges for governance and policy creation as new risks and capabilities emerge with further development in capabilities of AI Agents.
\end{itemize}
Future work should focus not only on enhancing agent capabilities but also on developing appropriate safeguards, ethical frameworks, and policies to guide implementation in real-world contexts.

\end{document}